\documentclass[conference]{IEEEtran}
\IEEEoverridecommandlockouts
\usepackage{cite}
\usepackage{amsmath,amssymb,amsfonts}
\usepackage{algorithmic}
\usepackage{graphicx}
\usepackage{multirow}
\usepackage{textcomp}
\usepackage{threeparttable}
\usepackage{arydshln}
\usepackage{xcolor}
\usepackage{subfigure}
\usepackage{fancyhdr}
\usepackage{ulem}
\def\BibTeX{{\rm B\kern-.05em{\sc i\kern-.025em b}\kern-.08em
    T\kern-.1667em\lower.7ex\hbox{E}\kern-.125emX}}
\begin{document}

\title{Biomedical named entity recognition using BERT in the machine reading comprehension framework
}

\author{\IEEEauthorblockN{Cong Sun\textsuperscript{1}, Zhihao Yang\textsuperscript{1,*}, Lei Wang\textsuperscript{2,*}, Yin Zhang\textsuperscript{2}, Hongfei Lin\textsuperscript{1}, Jian Wang\textsuperscript{1}}
\IEEEauthorblockA{\textit{\textsuperscript{1}School of Computer Science and Technology, Dalian University of Technology, Dalian, China, 116024} \\
\textit{\textsuperscript{2}Beijing Institute of Health Administration and Medical Information, Beijing, China, 100850}\\
\textsuperscript{*}Corresponding author: yangzh@dlut.edu.cn, wanglei.wlbihami@gmail.com}
}

\maketitle

\thispagestyle{fancy} 
\lhead{} 
\chead{} 
\rhead{} 
\lfoot{} 
\cfoot{} 
\rfoot{\thepage} 
\renewcommand{\headrulewidth}{0pt} 
\renewcommand{\footrulewidth}{0pt} 
      
\pagestyle{fancy}
\rfoot{\thepage}

\begin{abstract}
Recognition of biomedical entities from literature is a challenging research focus, which is the foundation for extracting a large amount of biomedical knowledge existing in unstructured texts into structured formats. Using the sequence labeling framework to implement biomedical named entity recognition (BioNER) is currently a conventional method. This method, however, often cannot take full advantage of the semantic information in the dataset, and the performance is not always satisfactory. In this work, instead of treating the BioNER task as a sequence labeling problem, we formulate it as a machine reading comprehension (MRC) problem. This formulation can introduce more prior knowledge utilizing well-designed queries, and no longer need decoding processes such as conditional random fields (CRF). We conduct experiments on six BioNER datasets, and the experimental results demonstrate the effectiveness of our method. Our method achieves state-of-the-art (SOTA) performance on the BC4CHEMD, BC5CDR-Chem, BC5CDR-Disease, NCBI-Disease, BC2GM and JNLPBA datasets, achieving F1-scores of 92.92\%, 94.19\%, 87.83\%, 90.04\%, 85.48\% and 78.93\%, respectively.
\end{abstract}


\section{Introduction}
Biomedical named entity recognition (BioNER) aims to automatically recognize biomedical entities (e.g., chemicals, diseases and proteins) in given texts. Effectively recognizing biomedical entities is the prerequisite for extracting biomedical knowledge deposited in unstructured texts, transforming them into structured formats. Therefore, the BioNER task has important research value. Traditionally, BioNER methods usually depend on well-designed feature engineering, i.e., the design of features uses various natural language processing (NLP) tools and domain knowledge. Typical representatives of such models used in the biomedical domain include DNorm\cite{leaman2013dnorm}, tmChem\cite{leaman2015tmchem}, TaggerOne\cite{leaman2016taggerone}, Lou's joint model\cite{lou2017transition}, etc. Feature engineering, however, relies heavily on domain-specific knowledge and hand-crafted features. Furthermore, these features are both model- and entity-specific. In recent years, neural networks with automatic feature learning abilities have become prevalent in NER tasks\cite{lample2016neural,jagannatha2016structured}. For the biomedical domain, several neural network methods\cite{habibi2017deep,dang2018d3ner,luo2018an,sachan2017effective,wang2018cross,yoon2019collabonet} have been proposed to recognize biomedical entities. Within these methods, bidirectional long short-term memory (BiLSTM)\cite{hochreiter1997long} is usually employed to learn vector representations of each word/token in a sentence, and then as the input to conditional random fields (CRF)\cite{lafferty2001conditional}. Very recently, language models (e.g., ELMo\cite{peters2018deep} and BERT\cite{devlin2019bert}) obtained state-of-the-art (SOTA) performance on many NLP tasks. In the biomedical domain, Lee et al.\cite{lee2019biobert} used BioBERT (namely BERT pre-trained on biomedical corpora) and the softmax function to recognize biomedical entities, and their method achieved SOTA results on several biomedical datasets. Compared with feature engineering methods, neural network methods are able to automatically learn features and thus can achieve more competitive performance.

The existing methods usually formalize the BioNER task into a sequence labeling problem, i.e., training a sequence labeling model to assign a label to each token in a given sequence. However, the models mentioned above, i.e., BiLSTM-CRF and BioBERT-Softmax, both cannot effectively learn the semantic information in the sequence labeling framework. For BiLSTM, its performance is unsatisfactory compared with language models (e.g., BioBERT)\cite{lee2019biobert}. For BioBERT, it is difficult to effectively use the semantic information learned by the final layer of BioBERT in the sequence labeling framework\cite{kaneko2019multi-head}. Inspired by the current trend of formalizing NLP tasks into machine reading comprehension (MRC) tasks\cite{levy2017zero-shot,mccann2018the,li2019entity-relation,shen2017reasonet,li-etal-2020-unified}, we use BioBERT in the MRC framework to perform BioNER (referred to BioBERT-MRC). In the MRC framework, each biomedical entity type can be encoded by a language query and identified by answering these queries. Take the sentence ``[Meloxicam]$_{chemical}$ - induced liver toxicity ." from the BC5CDR-Chem dataset as an example. We can introduce more prior knowledge (e.g., biomedical entities in the training/development datasets) by designing queries in the MRC framework. Therefore, the original sentence can be formalized as a sentence pair ``[Meloxicam]$_{chemical}$ - induced liver toxicity . \; Can you detect chemical entities like sodium or RA or cannabis ?", where ``sodium", ``RA" and ``cannabis" are chemical entities and can be obtained from the BC5CDR-Chem dataset. Compared with the sequence labeling framework, the MRC framework essentially has the advantage of introducing prior knowledge, which may contribute to improving model performance.

We use BioBERT in the MRC framework to perform BioNER tasks and conduct experiments on six biomedical datasets, i.e., BC4CHEMD\cite{krallinger2015the}, BC5CDR-Chem\cite{li2016BC5CDR}, BC5CDR-Disease\cite{li2016BC5CDR}, NCBI-Disease\cite{dogan2014ncbi}, BC2GM\cite{smith2008overview} and JNLPBA\cite{kim2004jnlpba} datasets. Our method achieves SOTA performance on all these datasets. The code and datasets can be found at https://github.com/CongSun-dlut/BioBERT-MRC.

\section{Related Work}
\subsection{Language Model}
Word embeddings can use a large amount of unlabeled data to learn the latent syntactic and semantic information of words/tokens, and map these words/tokens into dense low-dimensional vectors. In the past decade, several word-embedding methods have been proposed, among which the representative methods are Word2Vec\cite{mikolov2013distributed,mikolov2013efficient} and GloVe\cite{pennington2014glove}. Word2Vec employs the Skip-Gram model\cite{mikolov2013distributed} to predict surrounding words according to the current word/token or utilizes the Continuous Bag-Of-Words (CBOW) model\cite{mikolov2013efficient} to model the current word/token based on the surrounding context. GloVe\cite{pennington2014glove} uses a specific weighted least squares model that trains on global word-word co-occurrence counts, and thus effectively leverages statistics and both local and global features of the corpus. However, the word/token trained by these methods is mapped to a certain vector. Therefore, word embeddings trained by these methods can only model context-independent representations.

Nowadays, language models such as ELMo\cite{peters2018deep} and BERT\cite{devlin2019bert} do boost performance in NLP tasks. Unlike traditional word embeddings such as Word2Vec and GloVe, the embedding assigned to the word/token by the language model depends on the context, which means the same word/token could have different representations in different contexts. ELMo\cite{peters2018deep} leverages the concatenation of independently trained left-to-right and right-to-left LSTM to model the contextual information of the input sequence. BERT\cite{devlin2019bert} employs Transformer\cite{vaswani2017attention} to pre-train representations by jointly conditioning on both left and right context in all layers. Because the great success of BERT, it has gradually become a mainstream method using a large corpus to pre-train BERT and fine-tuning it on the target dataset.

\subsection{Machine Reading Comprehension}
The MRC methods\cite{levy2017zero-shot,mccann2018the,li2019entity-relation,shen2017reasonet,li-etal-2020-unified} could extract answer spans from the context through a given query. This task can be formalized as two classification tasks, namely to predict the start and end positions of the answer spans. In recent years, there has been a trend of converting related NLP tasks into MRC. For example, McCann et al.\cite{mccann2018the} used the question answering framework to implement ten different NLP tasks uniformly, and all achieved competitive performance. For the NER task, Li et al.\cite{li-etal-2020-unified} employed BERT to recognize entities from texts in the MRC framework (in the general domain). However, there is currently no specific research for BERT on BioNER in the MRC framework to the best of our knowledge. Our work focuses specifically on biomedical entities, which is significantly different from Li's work\cite{li-etal-2020-unified}. Furthermore, we are the first to explore the effect of different model components on BioNER tasks in the MRC framework.

\section{Methodology}
\subsection{Task Definition}
Given an input sentence $X$ = \{$x_1$, $x_2$, $\cdots$, $x_{N}$\}, where $x_i$ is the $i$-th word/token and $N$ represents the length of the sentence. The goal of NER is to classify each word/token in $X$ and assign it to a corresponding label $y$ $\in$ $Y$, where $Y$ is a predefined list of all possible label types (e.g., CHEMICAL, DISEASE and PROTEIN). We formulate the NER task as an MRC task and thus convert the labeling-style NER dataset into a set of (Context, Query, Answer) triples. Among them, Context is a given input sentence $X$, Query is a query sentence designed according to the sentence $X$, and Answer is the target entity span. For each label type $y$, we first constructed a query $Q_y$ = \{$q_1$, $q_2$, $\cdots$, $q_M$\} for each sentence, where $M$ represents the length of the query. Then, we obtained the annotated entities $x_{start,end}$ according to the annotated labels $Y$, where $x_{start,end}$ is a substring of $X$ and $start$ $\le$ $end$. For example, the sentence from the BC5CDR-Chem dataset, ``Meloxicam - induced liver toxicity .", its corresponding labels are ``B O O O O O". According to the labels, we can obtain the entities and their spans: ``Meloxicam"$_{0,0}$. Finally, we constructed the triple ($X$, $Q_y$, $x_{start,end}$), which is exactly the (Context, Query, Answer) triple that we need.

\subsection{Construct Queries}

Different from the query generation in the general domain, e.g., Li et al.\cite{li-etal-2020-unified} utilized the annotation guideline notes as references to construct queries, we used biomedical entities from the training/development set of the target dataset to construct queries. Table I lists some examples of the queries we constructed. We conducted experiments on three types of biomedical entities (i.e., CHEMICAL/DRUG, DISEASE and PROTEIN/GENE). The entity variables (such as $chemical_1$, $chemical_2$ and $chemical_3$) represent the annotated ones selected randomly from the training/development set of the target dataset.

\begin{table}[htb]
\caption{Examples of constructed queries.}
\centering
\begin{tabular}{p{1.8cm}p{6.0cm}}
\hline
Entity type & Query \\ \hline
CHEMICAL   & Can you detect chemical entities like $chemical_1$ or $chemical_2$ or $chemical_3$ ?\\
DISEASE    & Can you detect disease entities like $disease_1$ or $disease_2$ or $disease_3$ ?\\
PROTEIN    & Can you detect protein entities like $protein_1$ or $protein_2$ or $protein_3$ ?\\ \hline
\end{tabular}
\begin{tablenotes}
\item Notes: $chemical_x$, $disease_x$ and $protein_x$ are biomedical entities, which can be obtained from the training/development set of the target dataset.
\end{tablenotes}
\end{table}

\subsection{Model Details}

\begin{figure}[htb]
	\centering  
	\includegraphics[width=1.00\linewidth]{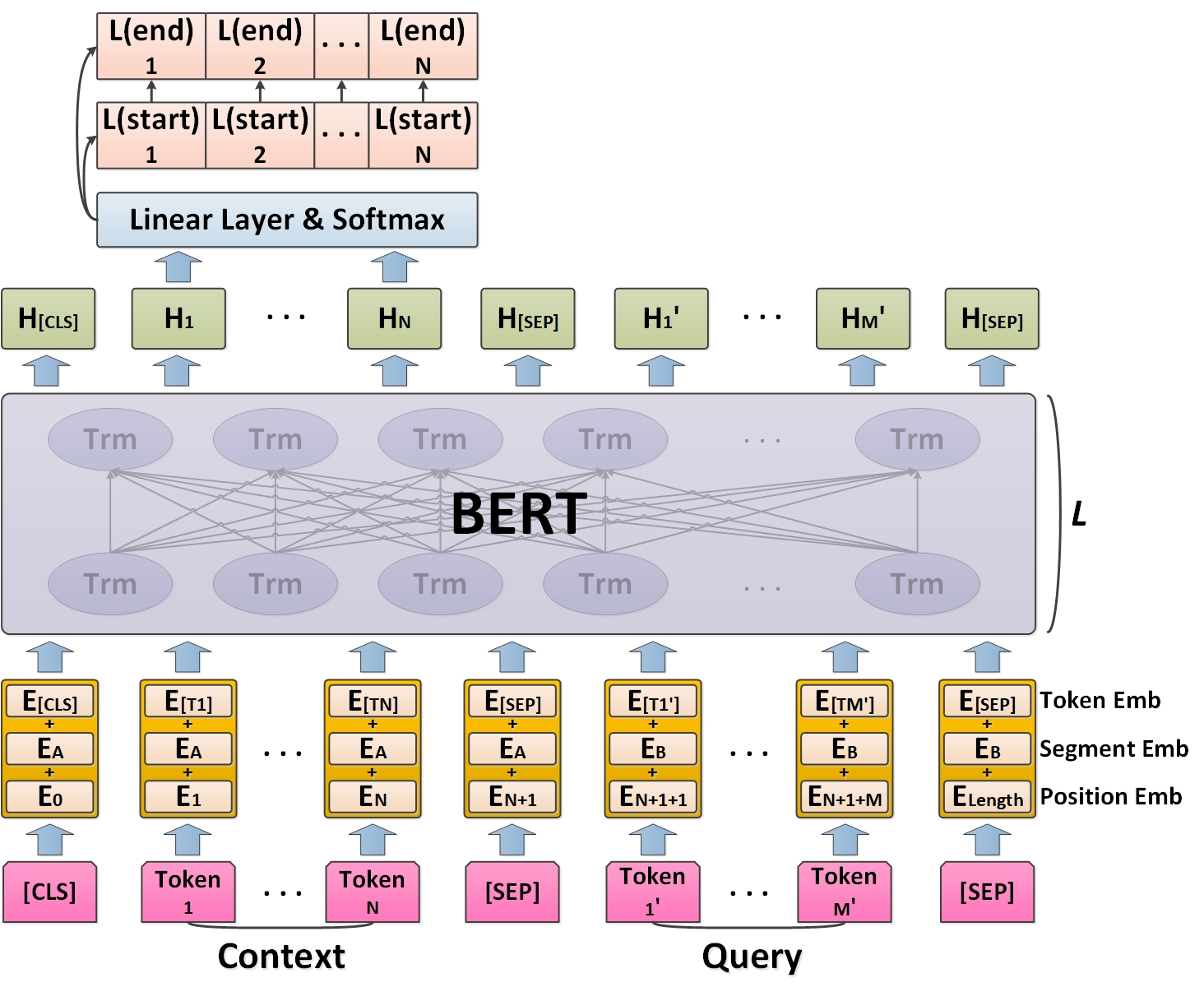}
	\caption{Using BERT to perform BioNER in the MRC framework.}  
\end{figure}

We exploited BERT\cite{devlin2019bert} as our model backbone. In this research, we need to recognize entities in the biomedical domain, so we employed BioBERT\cite{lee2019biobert} as the model weight. Figure 1 shows the BioNER task implemented in the MRC framework using BERT. Firstly, the query $Q_y$ is concatenated with the context $X$, forming the combined sequence \{[CLS], $x_1$, $x_2$, $\cdots$, $x_N$, [SEP], $q_1$, $q_2$, $\cdots$, $q_M$, [SEP]\}, where [CLS] and [SEP] are special tokens for marking the start and separation of tokens. Then, the combined sequence is fed into BERT, which can be defined by the following formulas:
\begin{equation}
h_i^0 = W_e  t_i + W_b
\end{equation}
\begin{equation}
h_i^l = Trm(h_i^{l-1})
\end{equation}
\begin{equation}
H = [h_1^L; h_2^L; \cdots; h_N^L] \in R^{N \times d}
\end{equation}
where $t_i$ is the embedding of the $i$-th token, $W_e$, $W_b$ are parameters, $L$ means the total number of layers for BERT, $l$ ($ 1 \leq l \leq L $) is the number for the current layer, and $Trm$ denotes the Transformer\cite{vaswani2017attention} block, including multi-head attention layers, fully connected layers and normalization layers. Furthermore, $H$ is the output of BERT and $N$ is the length of the context. We simply dropped query representations because they are not the target of the model prediction.

There are usually two strategies for choosing the span in the MRC framework. The first is to employ two $n$-class classifiers to respectively predict the start and end indexes, where $n$ is the length of the context. Because the function is calculated on all tokens in the entire context, this strategy has the deficiency that one input sequence can only output one span. The other is to construct two binary classifiers. One is used to predict whether the token is a start index, and the other serves to predict whether the token is an end index. This strategy allows multiple start and end indexes to be output for a given sequence, and therefore it has the potentials to identify all target entities based on $Q_y$. 

In this study, we adopted the second strategy. Given the representation matrix $H$ output by BERT, the model first predicts the probability of each token being a start index. The formula is as follows:
\begin{equation}
L_{start} = linear (H  W_{start}) \; \in R^{N\times2}
\end{equation}
where $linear$ is a fully connected layer and $W_{start}$ is the weight to learn. Each row of $L_{start}$ denotes the hidden representation of the index, used to determine the starting position of a target entity for a given query. 

Then, the model predicts the probability of each token being the corresponding end index. The formula is:

\begin{equation}
L_{end} = linear (H  W_{end};softmax(L_{start})) \; \in R^{N\times2}
\end{equation}
where $linear$ is a fully connected layer, $W_{end}$ is the weight to learn, and ``;" means the concatenation operation.

Finally, by applying the argmax function to each row of $L_{start}$ and $L_{end}$, we can obtain the predicted indexes that might be the start or end positions, i.e., $I_{start}$ and $I_{end}$:
\begin{equation}
I_{start} = \{i \, | \, argmax(L_{start}^i) = 1, \;\;i = 1, 2, \cdots, N\}
\end{equation}

\begin{equation}
I_{end} = \{j \, | \, argmax(L_{end}^j) = 1, \;\;j = 1, 2, \cdots, N\}
\end{equation}
where the superscript $i$ and $j$ denote the $i$-th and $j$-th rows of a matrix, respectively.

In context $X$, there could be multiple entities of the same entity type. This means that multiple start/end indexes can be predicted from one input sequence at one time. Because the datasets we conducted experiments on are all flat NER datasets, we further used the nearest match principle to match the start and end indexes to obtain final answers. Specifically, for the case where one end/start index corresponds to multiple start/end indexes, only the nearest start/end index is matched.

\subsection{Training and Test}

At the training time, context $X$ is paired with two label sequences $Y_{start}$ and $Y_{end}$ of length $N$ representing the ground-truth label of each token $x_i$ being the start/end index of the biomedical entity. The loss function is defined as follows:

\begin{equation}
Loss_{start} = CE(L_{start}, Y_{start})
\end{equation}

\begin{equation}
Loss_{end} = CE(L_{end}, Y_{end})
\end{equation}

\begin{equation}
Loss = (Loss_{start} + Loss_{end})/2
\end{equation}
where $CE$ denotes the cross-entropy loss function.

At the test time, the start and end indexes are first separately selected based on $I_{start}$ and $I_{end}$. Then, the nearest match principle is used to match the start and end indexes to obtain the final answers. 

\section{Experiments}
\subsection{Datasets and Experimental Settings}

Our method is evaluated on BC4CHEMD\cite{krallinger2015the}, BC5CDR\cite{li2016BC5CDR}, NCBI-Disease\cite{dogan2014ncbi}, BC2GM\cite{smith2008overview} and JNLPBA\cite{kim2004jnlpba} datasets, all of which are pre-processed and provided by Lee et al.\cite{lee2019biobert}. Among these datasets, BC5CDR has two sub-datasets, BC5CDR-Chem and BC5CDR-Disease, which are used to evaluate chemical and disease entities, respectively. Because most of the existing methods were evaluated on BC5CDR-Chem and BC5CDR-Disease respectively, we did the same. Table II lists the statistics of these datasets.

\begin{table}[htb]
\centering
\caption{Statistics of BioNER datasets.}
\begin{tabular}{p{2.07cm}p{1.7cm}p{1.78cm}p{1.56cm}}
\hline
Dataset         & Entity type      & No. annotations  &No. sentences \\ \hline
BC4CHEMD        & Chemical/Drug       & 79,842        &89,679 \\
BC5CDR-Chem     & Chemical/Drug       & 15,411        &14,228 \\
BC5CDR-Disease  & Disease             & 12,694        &14,228 \\
NCBI-Disease    & Disease             & 6,881         &7,639 \\
BC2GM           & Protein/Gene        & 20,703        &20,510 \\
JNLPBA          & Protein/Gene        & 35,460        &22,562 \\
\hline   
\end{tabular}
\begin{tablenotes}
\item Notes: Statistics are from Habibi et al.\cite{habibi2017deep}.
\end{tablenotes}
\end{table}

In the experiments, the original training and development set were merged into a new training set. Then 10\% of the new training set was sampled as the validation set to tune hyper-parameters. The test set was only used to evaluate the model. Most existing works\cite{leaman2016taggerone,luo2018an,sachan2017effective,wang2018cross,yoon2019collabonet,lee2019biobert} split data in this way, and we also followed this way. Because the limitation of computational complexity, most of the existing works are based on BERT$_{\!B\!A\!S\!E}$ \cite{devlin2019bert}. To facilitate comparison with these works, all BERT models in this work are based on the BERT$_{\!B\!A\!S\!E}$ model. The performance is measured with the F1-score ($F1$), whose attributes equal importance to precision ($P$) and recall ($R$). The formula is: $F1$ = 2$PR$/($P$+$R$). In this work, each experiment is repeated five times, and we report the maximum F1-score (referred to max), average F1-score (referred to mean), and standard deviation (referred to std). Moreover, we also exploit T-TEST to perform statistical significance tests and report the confidence interval. Table III lists the detailed hyper-parameters of BioBERT-MRC. In our experiments, BioBERT-MRC reaches its highest performance at 1 or 3 epochs on the BC4CHEMD, BC5CDR-Disease, BC2GM and JNLPBA datasets. The reasons are two-fold: 1) these datasets, especially BC4CHEMD, are large in scale; and 2) BioBERT-MRC has powerful feature learning capabilities. Furthermore, note that the input of BioBERT-MRC is ``[CLS] Context [SEP] Query [SEP]" instead of ``[CLS] Query [SEP] Context [SEP]". We also explored using ``[CLS] Query [SEP] Context [SEP]" as input, but the model performance has not been further improved. The detailed discussions are provided in Supplementary Material A.1.

\begin{table}[htb]
\centering
\caption{The detailed hyper-parameters of BioBERT-MRC.}
\begin{tabular}{p{2.3cm}p{0.9cm}p{0.9cm}p{0.65cm}p{0.65cm}p{0.65cm}}
\hline
Dataset           & seq\_len    & epochs  & bs  & lr     & loss \\ \hline
BC4CHEMD          & 512         & 1       & 8   & 3e-5   & CE \\
BC5CDR-Chem       & 256         & 10      & 16  & 3e-5   & CE \\
BC5CDR-Disease    & 256         & 3       & 16  & 3e-5   & CE \\
NCBI-Disease      & 256         & 15      & 16  & 3e-5   & CE \\
BC2GM             & 256         & 3       & 16  & 3e-5   & CE \\
JNLPBA            & 256         & 3       & 16  & 3e-5   & CE \\
\hline   
\end{tabular}
\begin{tablenotes}
\item Notes: ``seq\_len" denotes the maximum sequence length, ``bs" denotes the batch size, ``lr" denotes the learning rate, and ``CE" means the cross-entropy loss function. Moreover, ``epochs" refers to the experimental setting when BioBERT-MRC obtains SOTA performance.
\end{tablenotes}
\end{table}

\subsection{The Effect of Different BERTs on Performance}
To explore the effect of different BERTs on BERT-MRC, we compared three remarkable BERT models in the biomedical domain, i.e., BioBERT\cite{lee2019biobert}, BlueBERT\cite{peng2019transfer} and ClinicalBERT\cite{alsentzer2019publicly}. Specifically, we chose BioBERTv1.1 (+PubMed, Cased)\cite{lee2019biobert}, BlueBERT (+PubMed, UnCased)\cite{peng2019transfer} and Bio-Discharge-Summary-BERT (Cased)\cite{alsentzer2019publicly} as the representatives. These models have obtained SOTA performance in their works, respectively. Table IV illustrates the effect of different BERTs on model performance. Overall, the performance of BioBERT-MRC is better than BlueBERT-MRC and ClinicalBERT-MRC. Compared with BlueBERT, BioBERT is sensitive to uppercase and lowercase characters. This experimental result shows that the uppercase and lowercase character information are useful for BioNER tasks on most datasets. Moreover, we also noticed that the performance of BlueBERT-MRC (94.11\% in the average F1-score) is superior to BioBERT-MRC (93.92\% in the average F1-score) on the BC5CDR-Chem dataset. We observed from the corpus that most of the target entities in the test set are lowercase entities, which is one reason BlueBERT-MRC performs better on this dataset. On the other hand, BioBERT concentrates on PubMed abstracts to perform pre-training, while ClinicalBERT uses PubMed abstracts, PMC full-text articles, and Discharge Summaries for pre-training. The experimental result indicates that the pre-training corpus greatly impacts performance, and BioBERT pre-trained on the PubMed abstracts is more suitable for the BioNER tasks in our experiments.

\begin{table}[tb]
\centering
\caption{Performance comparison for different BERTs.}
\begin{tabular}{p{2.25cm}p{2.55cm}p{1.6cm}p{0.6cm}}
\hline
\textbf{Dataset}     &\textbf{Model}   &\textbf{mean$\pm$std}  &\textbf{max}\\ \hline
BC4CHEMD    &ClinicalBERT-MRC     &91.28$\pm$0.06  &91.35 \\ 
            &BlueBERT-MRC       	&91.87$\pm$0.18   &91.99\\
            &BioBERT-MRC  &\textbf{92.70$\pm$0.16}   &\textbf{92.92}\\
\cdashline{1-4}[2pt/2pt]
BC5CDR-Chem    &ClinicalBERT-MRC     &93.29$\pm$0.10   & 93.34\\ 
            &BlueBERT-MRC            	&\textbf{94.11$\pm$0.10}  &\textbf{94.19}\\
            &BioBERT-MRC &93.92$\pm$0.18   &\textbf{94.19}\\
\cdashline{1-4}[2pt/2pt]
BC5CDR-Disease   &ClinicalBERT-MRC    &86.36$\pm$0.20   & 86.67\\
            &BlueBERT-MRC     &87.53$\pm$0.22   &\textbf{87.84}\\
            &BioBERT-MRC &\textbf{87.56$\pm$0.19}  &87.83\\
\cdashline{1-4}[2pt/2pt]            
NCBI-Disease   &ClinicalBERT-MRC  	&88.17$\pm$0.40   &88.82\\ 
            &BlueBERT-MRC     	&88.29$\pm$0.59	 &88.82\\
            &BioBERT-MRC    &\textbf{89.39$\pm$0.38}  &\textbf{90.04}\\
\cdashline{1-4}[2pt/2pt]
BC2GM    &ClinicalBERT-MRC    	&84.11$\pm$0.15   &84.37\\ 
            &BlueBERT-MRC    &84.85$\pm$0.09   &84.93\\
            &BioBERT-MRC &\textbf{85.11$\pm$0.39}   &\textbf{85.48}\\
\cdashline{1-4}[2pt/2pt]            
JNLPBA   &ClinicalBERT-MRC   &77.64$\pm$0.14   &77.86\\ 
            &BlueBERT-MRC    	&78.29$\pm$0.20  &78.59\\
            &BioBERT-MRC  	&\textbf{78.45$\pm$0.29}  &\textbf{78.93}\\
\hline   
\end{tabular}
\begin{tablenotes}
\item Notes: ``max" denotes the maximum F1-score, ``mean" denotes the average F1-score, and ``std" denotes the standard deviation. The best scores are shown in bold.
\end{tablenotes}
\end{table}

\subsection{MRC vs Sequence Labeling}

\begin{table}[tb]
\centering
\caption{Performance comparison for different models.}
\begin{tabular}{p{2.25cm}p{3.15cm}l}
\hline
\textbf{Dataset}     &\textbf{Model}   &\textbf{mean$\pm$std (\%)}\\ \hline
BC4CHEMD    &BioBERT-Softmax &91.99$\pm$0.19\\
            &BioBERT-CRF &91.52$\pm$0.31\\
            &BioBERT-BiLSTM-CRF  &91.39$\pm$0.37\\
            &BioBERT-MRC  &\textbf{92.70$\pm$0.16}\\
\cdashline{1-3}[2pt/2pt]
BC5CDR-Chem    &BioBERT-Softmax  &93.65$\pm$0.04\\
            &BioBERT-CRF &93.62$\pm$0.07\\
            &BioBERT-BiLSTM-CRF  &93.43$\pm$0.24\\
            &BioBERT-MRC  &\textbf{93.92$\pm$0.18}\\
\cdashline{1-3}[2pt/2pt]
BC5CDR-Disease   &BioBERT-Softmax  &86.60$\pm$0.18\\
            &BioBERT-CRF &86.25$\pm$0.39\\
            &BioBERT-BiLSTM-CRF  &86.16$\pm$0.44\\
            &BioBERT-MRC  &\textbf{87.56$\pm$0.19}\\
\cdashline{1-3}[2pt/2pt]            
NCBI-Disease   &BioBERT-Softmax  &88.36$\pm$0.50\\ 
            &BioBERT-CRF &88.60$\pm$0.57\\
            &BioBERT-BiLSTM-CRF  &88.62$\pm$0.40\\
            &BioBERT-MRC  &\textbf{89.39$\pm$0.38}\\
\cdashline{1-3}[2pt/2pt]
BC2GM    &BioBERT-Softmax  &84.47$\pm$0.20\\
            &BioBERT-CRF&84.44$\pm$0.18\\
            &BioBERT-BiLSTM-CRF  &84.38$\pm$0.14\\
            &BioBERT-MRC  &\textbf{85.11$\pm$0.39}\\
\cdashline{1-3}[2pt/2pt]            
JNLPBA   &BioBERT-Softmax &77.80$\pm$0.47\\
            &BioBERT-CRF&77.55$\pm$0.53\\
            &BioBERT-BiLSTM-CRF &77.61$\pm$0.51\\
            &BioBERT-MRC  &\textbf{78.45$\pm$0.29}\\
\hline   
\end{tabular}
\begin{tablenotes}
\item Notes: ``mean" denotes the average F1-score, and ``std" denotes the standard deviation. The best scores are shown in bold.
\end{tablenotes}
\end{table}

In this work, we explored the effect of BERT in the MRC framework and the sequence labeling framework in detail. First, we compared the performance of BERT on BioNER in the MRC framework and the sequence labeling framework. Table V shows the performance comparison for different models. BioBERT-Softmax, BioBERT-CRF and BioBERT-BiLSTM-CRF are common methods used in the sequence labeling framework. For these methods, BioNER can be deemed a problem of token-level sequence labeling, and each word in the input sequence is assigned to a label with the BIO (i.e., Begin, Inside and Outside) tagging scheme. The goal is to predict the BIO label of each token in the output sequence. BioBERT-Softmax uses the softmax function to perform the BIO classification on each token's BioBERT output in the sequence. BioBERT-CRF employs CRFs to calculate each token's BioBERT output in the sequence to obtain the best sequence labeling result. BioBERT-BiLSTM-CRF utilizes the BiLSTM-CRF model to predict each token's BioBERT output in the sequence. Different from these three methods, BioBERT-MRC views BioNER as a machine reading comprehension problem and predicts answer spans $x_{start,end}$ based on the input sequence $X$ and the query $Q_y$. As listed in Table V, the performance of the three sequence labeling methods (i.e., BioBERT-Softmax, BioBERT-CRF and BioBERT-BiLSTM-CRF) is similar. On the BC4CHEMD dataset, the performance of BioBERT-Softmax (91.99\% in the average F1-score) is superior to BioBERT-CRF (91.52\% in the average F1-score) and BioBERT-BiLSTM-CRF (91.39\% in the average F1-score). The reason may be that the sentences in the dataset are relatively long, while CRF and LSTM seem not good at learning long-distance hidden representations. However, the performance of these three methods on all six datasets is inferior to BioBERT-MRC. Compared with the sequence labeling methods, BioBERT-MRC can greatly improve the performance of BioNER tasks regardless of the corpus scale and sentence length. These experimental results show that BERT's capability of recognizing biomedical entities in the MRC framework is superior to its capability of recognizing biomedical entities in the sequence labeling framework.

\subsection{The Effect of Different MRC Strategies on Performance}
In this work, we compared the effect of different MRC strategies on model performance. First, we explored the effect of different span strategies. As described in Section III.C $Model \; Details$, BioBERT-MRC uses the start\_index information in Equation 5. In this section, we discuss the effect of start\_index information on BioBERT-MRC. Specifically, we designed a baseline model, BioBERT-MRC$^a$, and compared the differences between BioBERT-MRC$^a$ and BioBERT-MRC. In terms of implementation, BioBERT-MRC$^a$ only replaces Equation 5 of BioBERT-MRC (described in Section III.C) with the following equation, while the rest remain unchanged:
\begin{equation}
L_{end} = linear (H  W_{end}) \; \in R^{N\times2}
\end{equation} 
Table VI shows the performance comparison of these two models. It can be seen that the average F1-scores of these two models are both competitive. Overall, BioBERT-MRC obtains better performance on five of the six datasets. Compared with other types of biomedical entities, we observed that disease entities are more sensitive to the introduction of start\_index information. Compared with BioBERT-MRC$^a$, the average F1-scores of BioBERT-MRC on the BC5CDR-Disease and NCBI-Disease datasets increased by 0.41\% and 0.59\%, respectively. This experimental result indicates that different end\_index functions can affect model performance to a certain extent, and considering the start index may improve performance when predicting the end index.

\begin{table}[tb]
\centering
\caption{Performance comparison for different end index strategies.}
\begin{tabular}{p{2.5cm}p{2.6cm}l}
\hline
\textbf{Dataset}     &\textbf{Model}   &\textbf{mean$\pm$std (\%)} \\\hline
BC4CHEMD    
            &BioBERT-MRC$^a$  &92.60$\pm$0.08\\
            &BioBERT-MRC  &\textbf{92.70$\pm$0.16}\\
\cdashline{1-3}[2pt/2pt]
BC5CDR-Chem    
            &BioBERT-MRC$^a$ &93.86$\pm$0.17\\
            &BioBERT-MRC  &\textbf{93.92$\pm$0.18}\\  
\cdashline{1-3}[2pt/2pt]
BC5CDR-Disease   
            &BioBERT-MRC$^a$ &87.15$\pm$0.13\\
            &BioBERT-MRC &\textbf{87.56$\pm$0.19}\\ 
\cdashline{1-3}[2pt/2pt]            
NCBI-Disease   
            &BioBERT-MRC$^a$ &88.80$\pm$0.47\\
            &BioBERT-MRC &\textbf{89.39$\pm$0.38}\\ 
\cdashline{1-3}[2pt/2pt]
BC2GM    
            &BioBERT-MRC$^a$  &\textbf{85.14$\pm$0.12}\\
            &BioBERT-MRC &85.11$\pm$0.39\\ 
\cdashline{1-3}[2pt/2pt]            
JNLPBA   
            &BioBERT-MRC$^a$ &78.27$\pm$0.15\\
            &BioBERT-MRC&\textbf{78.45$\pm$0.29}\\
\hline   
\end{tabular}
\begin{tablenotes}
\item Notes: ``mean" denotes the average F1-score, and ``std" denotes the standard deviation. The best scores are shown in bold.
\end{tablenotes}
\end{table}

Then, we explored the effect of constructed queries. Intuitively, the more information a query contains, the better model performance should be. However, if too many biomedical entities were introduced into Query, it may cause the model to be more biased towards the learning for introduced ``knowledge", leading to an ``overfitting" problem. Therefore, how many biomedical entities are introduced to construct Query to obtain competitive performance is a meaningful design indicator. In the experiments, in addition to the constructed queries (referred to Query-3) described in Section III.B $Construct \; Queries$, we further designed four types of queries (i.e., Query-5, Query-10, Query-0 and Query-None) to explore the effect of Query:

\begin{itemize}
\item
\textbf{Query-5}: Similar to Query-3, except that it randomly select five entities from the training/development set.
\item
\textbf{Query-10}: Similar to Query-3, except that it randomly select ten entities from the training/development set.

\item 
\textbf{Query-0}:  Simple language query, i.e., ``Can you detect $X$-$type$ entities ?". For example, for the CHEMICAL entity, the query is ``Can you detect chemical entities ?".
\item
\textbf{Query-None}: The query is designed as ``none".
\end{itemize}

\begin{figure}[tbp]
	\centering  
	\includegraphics[width=1.00\linewidth]{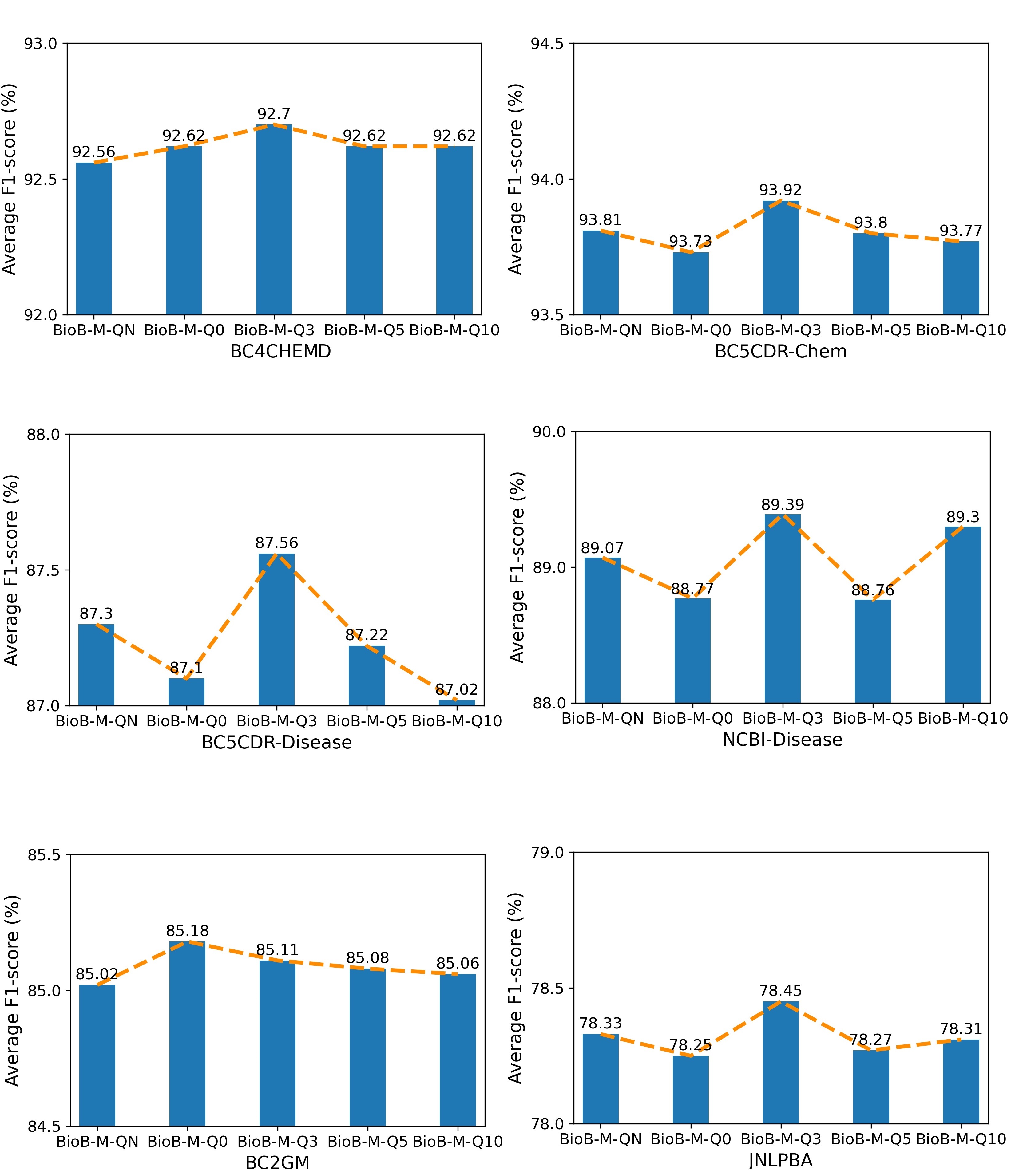}
	\caption{The performance comparison for different types of queries. BioB-M-QN, BioB-M-Q0, BioB-M-Q3, BioB-M-Q5 and BioB-M-Q10 denote BioBERT-MRC (Query-None), BioBERT-MRC (Query-0), BioBERT-MRC (Query-3), BioBERT-MRC (Query-5) and BioBERT-MRC (Query-10), respectively.} 
\end{figure}

In this way, we can comprehensively analyze the effect of Query on BioBERT-MRC. Figure 2 illustrates the performance comparison for different types of queries. BioB-M-QN, BioB-M-Q0, BioB-M-Q3, BioB-M-Q5 and BioB-M-Q10 denote BioBERT-MRC (Query-None), BioBERT-MRC (Query-0), BioBERT-MRC (Query-3), BioBERT-MRC (Query-5) and BioBERT-MRC (Query-10), respectively. It can be seen that different queries have a great impact on the performance of BioBERT-MRC. Overall, the performance of BioBERT-MRC has a curve like ``$\frown$" on each dataset, and BioBERT-MRC (Query-3) obtains SOTA performance on five of these six datasets. We noticed that BioBERT-MRC (Query-None) and BioBERT-MRC (Query-0) also obtained competitive performance on these six datasets. This shows that even if the external knowledge introduced by Query is restricted, BioBERT can still utilize the MRC framework to obtain competitive performance. Specifically, the performance of BioBERT-MRC constructed using different queries has its own characteristics on each dataset. We were surprised to find that BioBERT-MRC (Query-0) obtain SOTA performance on the BC2GM dataset. The reason could be that BioBERT-MRC is not sensitive to external knowledge introduced by Query on the BC2GM dataset.

\begin{table*}[htbp]
\caption{Performance comparison with other existing methods.}
\begin{center}
\begin{tabular}{p{2.67cm}p{0.3cm}p{0.3cm}p{0.59cm}p{0.3cm}p{0.3cm}p{0.59cm}p{0.3cm}p{0.3cm}p{0.59cm}p{0.3cm}p{0.3cm}p{0.59cm}p{0.3cm}p{0.3cm}p{0.59cm}p{0.3cm}p{0.3cm}p{0.7cm}}
\hline
\multirow{2}{*}{\textbf{Method}}&\multicolumn{3}{c}{\textbf{BC4CHEMD}}&\multicolumn{3}{c}{\textbf{BC5CDR-Chem}}&\multicolumn{3}{c}{\textbf{BC5CDR-Disease}}&\multicolumn{3}{c}{\textbf{NCBI-Disease}}&\multicolumn{3}{c}{\textbf{BC2GM}}&\multicolumn{3}{c}{\textbf{JNLPBA}} \\
\cline{2-19} 
&\textbf{\textit{P(\%)}}&\textbf{\textit{R(\%)}}&\textbf{\textit{F1(\%)}} &\textbf{\textit{P(\%)}}&\textbf{\textit{R(\%)}}&\textbf{\textit{F1(\%)}} &\textbf{\textit{P(\%)}}&\textbf{\textit{R(\%)}}&\textbf{\textit{F1(\%)}} &\textbf{\textit{P(\%)}}&\textbf{\textit{R(\%)}}&\textbf{\textit{F1(\%)}} &\textbf{\textit{P(\%)}}&\textbf{\textit{R(\%)}}&\textbf{\textit{F1(\%)}} &\textbf{\textit{P(\%)}}&\textbf{\textit{R(\%)}}&\textbf{\textit{F1(\%)}}\\
\hline
tmChem\cite{leaman2015tmchem} &89.09&85.75&87.39$^{**}$ &--&--&-- &--&--&-- &--&--&-- &--&--&-- &--&--&--\\
TaggerOne\cite{leaman2016taggerone} &--&--&-- &94.20&88.80&91.40$^{**}$ &85.20&80.20&82.60$^{**}$ &85.10&80.80&82.90$^{**}$ &--&--&-- &--&--&--\\
Lou et al.\cite{lou2017transition}&--&--&-- &--&--&-- &89.61&83.09&86.23$^{**}$ &90.72&74.89&82.05$^{**}$ &--&--&-- &--&--&-- \\
D3NER\cite{dang2018d3ner} &--&--&-- &93.73&92.56&93.14$^{**}$ &83.98&85.40&84.68$^{**}$ &85.03&83.80&84.41$^{**}$ &--&--&-- &--&--&--\\
Luo et al.\cite{luo2018an} &92.29&90.01&91.14$^{**}$ &93.49&91.68&92.57$^{**}$ &--&--&-- &--&--&-- &--&--&-- &--&--&--\\
Sachan et al.\cite{sachan2017effective} &--&--&-- &--&--&-- &--&--&-- &86.41&88.31&87.34$^{**}$ &81.81&81.57&81.69$^{**}$ &71.39&79.06&75.03$^{**}$\\
Wang et al.\cite{wang2018cross} &91.30&87.53&89.37$^{**}$ &--&--&-- &--&--&-- &85.86&86.42&86.14$^{**}$ &82.10&79.42&80.74$^{**}$ &70.91&76.34&73.52$^{**}$\\
CollaboNet\cite{yoon2019collabonet} &90.78&87.01&88.85$^{**}$ &94.26&92.38&93.31$^{**}$ &85.61&82.61&84.08$^{**}$ &85.48&87.27&86.36$^{**}$ &80.49&78.99&79.73$^{**}$ &74.43&83.22&\underline{78.58}\\
BioBERT-Softmax\cite{lee2019biobert} &92.80&91.92&\underline{92.36}$^*$ &93.68&93.26&\underline{93.47}$^{*}$  &86.47&87.84&\underline{87.15}$^{**}$ &88.22&91.25&\underline{89.71}$^{*}$ &84.32&85.12&\underline{84.72}$^{*}$ &72.24&83.56&77.49$^{*}$\\
BERT-MRC\cite{li-etal-2020-unified} (L) &91.90&88.88&90.36$^{**}$ &92.98&93.72&93.35$^{**}$ &86.43&87.25&86.84$^{**}$ &88.16&90.00&89.07$^{**}$ &83.27&84.70&83.98$^{**}$ &72.35&84.55&77.98$^{*}$\\
BioBERT-MRC &93.89 &91.96&\textbf{92.92} &94.37 &94.00 &\textbf{94.19} &88.61 &87.07 &\textbf{87.83} &89.67&90.42&\textbf{90.04} &87.04&83.98&\textbf{85.48} &75.96&82.13&\textbf{78.93}\\
\hline
\end{tabular}
\end{center}
\begin{tablenotes}
\item  Notes: The first part (i.e., rows 1–3) is feature engineering-based methods, the second part (i.e., rows 4-8) is LSTM-CRF-based methods, and the third part (i.e., row 9–11) is BERT-based methods. BERT-MRC(L) denotes BERT-MRC run locally on the six BioNER datasets, with the same weights and queries as BioBERT-MRC. BioBERT-Softmax, BERT-MRC and BioBERT-MRC are all based on BioBERT v1.1 (+PubMed)\cite{lee2019biobert}. $^*$ and $^{**}$ denote a significant difference between the means of two models according to the T-TEST statistical test. Specifically, $^*$ indicates the model has a significant difference compared with BioBERT-MRC, with a more than 95\% confidence interval ($p$ \textless 0.05); $^{**}$ indicates the model has a significant difference compared with BioBERT-MRC, with a more than 99\% confidence interval ($p$ \textless 0.01). The best scores are shown in bold, and the second-best scores are underlined.
\end{tablenotes}
\end{table*}

\subsection{Performance Comparison with Other Methods}
Table VII shows the experimental results on BioNER datasets. The first three methods all rely on feature engineering. TmChem\cite{leaman2015tmchem} is a method that utilizes chemical knowledge to design features to recognize chemical entities. TaggerOne\cite{leaman2016taggerone} and Lou's model\cite{lou2017transition} both perform BioNER in a joint manner with named entity normalization (NEN), in which the feedback from NEN can be used to reduce NER errors. The difference is the former uses a semi-Markov linear classifier for BioNER, while the latter is based on a finite state machine. However, these methods are often model- and entity-specific and the performance is unsatisfied. In addition to the first three methods, the others are all neural network-based methods. Among these methods, the first five mainly use the LSTM-CRF model, while the latter three leverage the BERT model. D3NER\cite{dang2018d3ner} introduces various linguistic information to improve performance. Luo et al.\cite{luo2018an} utilized document-level global information obtained by the attention mechanism to enforce tagging consistency across multiple instances of the same token in the document. Sachan et al.\cite{sachan2017effective} trained a bidirectional language model on unlabeled data and transfer the weights to pre-train a BioNER model. Wang et al.\cite{wang2018cross} proposed a multi-task learning framework for BioNER to collectively use the training data of different types of entities. CollaboNet\cite{yoon2019collabonet} can also be regarded as a multi-task learning framework, which improves performance by combining multiple BioNER models. Generally, the performance of these LSTM-CRF-based models is superior to feature engineering-based models.

\begin{table*}[htbp]
\centering
\caption{Representative results of case study.}
\begin{tabular}{llllp{8.25cm}}
\hline
\textbf{No.} &\textbf{Dataset} &\textbf{Gold standard} &\textbf{Model} &\textbf{Result}\\ \hline
\multirow{4}{*}{1} &\multirow{4}{*}{BC4CHEMD} &\multirow{4}{2.5cm}{\textcolor{teal}{lipoic acid (CHEMICAL)}}   &\multirow{2}{*}{BioBERT-Softmax}      &$\cdots$ and the natural antioxidant lipoic acid , without influencing the level of free $\cdots$\\
                            &&&\multirow{2}{*}{BioBERT-MRC}          &$\cdots$ and the natural antioxidant \textcolor{teal}{lipoic acid} , without influencing the level of free $\cdots$\\
                            \cdashline{1-5}[2pt/2pt]
\multirow{4}{*}{2} &\multirow{4}{*}{BC5CDR-Chem}  &\multirow{4}{2.7cm}{\textcolor{teal}{cocaine (CHEMICAL)} DSE (OTHER)}  &\multirow{2}{*}{BioBERT-Softmax}      &$\cdots$ cocaine use , we conducted a pilot study to assess the safety of \textcolor{teal}{DSE} in emergency department patients with \textcolor{teal}{cocaine} - associated chest pain .\\
                            &&&\multirow{2}{*}{BioBERT-MRC}          &$\cdots$ \textcolor{teal}{cocaine} use , we conducted a pilot study to assess the safety of DSE in emergency department patients with \textcolor{teal}{cocaine} - associated chest pain . \\
                            \cdashline{1-5}[2pt/2pt]
\multirow{4}{*}{3} &\multirow{4}{*}{BC5CDR-Disease}  &\multirow{4}{2.6cm}{drug - induced disease (OTHER)}   &\multirow{2}{*}{BioBERT-Softmax}      &Differential diagnosis for \textcolor{cyan}{drug - induced disease} is invaluable even for patients $\cdots$\\
                            &&&\multirow{2}{*}{BioBERT-MRC}           &Differential diagnosis for drug - induced disease is invaluable even for patients $\cdots$\\
                            \cdashline{1-5}[2pt/2pt]
\multirow{4}{*}{4} &\multirow{4}{*}{NCBI Disease} &\multirow{4}{2.4cm}{\textcolor{cyan}{dominantly inherited neurodegeneration (DISEASE)}}    &\multirow{2}{*}{BioBERT-Softmax}      &This mutation may be valuable for developing models of dominantly inherited \textcolor{cyan}{neurodegeneration} $\cdots$\\
                            &&&\multirow{2}{*}{BioBERT-MRC}           &This mutation may be valuable for developing models of \textcolor{cyan}{dominantly inherited neurodegeneration} $\cdots$ \\
                            \cdashline{1-5}[2pt/2pt]
\multirow{4}{*}{5} &\multirow{4}{*}{BC2GM}  &\multirow{4}{2.4cm}{\textcolor{orange}{RXR ligand binding domain (PROTEIN)}}   &\multirow{2}{*}{BioBERT-Softmax}      &$\cdots$ influences the \textcolor{orange}{RXR} ligand binding domain such that it is resistant to the binding of 9 - cis RA $\cdots$\\
                            &&&\multirow{2}{*}{BioBERT-MRC}           &$\cdots$ influences the \textcolor{orange}{RXR ligand binding domain} such that it is resistant to the binding of 9 - cis RA $\cdots$\\
                            \cdashline{1-5}[2pt/2pt]
\multirow{4}{*}{6} &\multirow{4}{*}{JNLPBA}  &\multirow{4}{2.95cm}{peroxisome proliferator - activated receptor gamma (OTHER)}   &\multirow{2}{*}{BioBERT-Softmax}      &Oxidized alkyl phospholipids are specific , high affinity \textcolor{orange}{peroxisome proliferator - activated receptor gamma} ligands and agonists .\\
                            &&&\multirow{2}{*}{BioBERT-MRC}           &Oxidized alkyl phospholipids are specific , high affinity peroxisome proliferator - activated receptor gamma ligands and agonists .\\
\hline   
\end{tabular}
\begin{tablenotes}
\item Notes: Green represents chemical entities, blue represents disease entities, and yellow represents protein entities.
\end{tablenotes}
\end{table*}

Most recently, Lee et al.\cite{lee2019biobert} utilized BioBERT to perform BioNER in the sequence labeling framework (BIO tagging scheme), which raises the performance of BioNER to a new level. Existence does not entail exclusivity. Although the use of sequence tagging for BioNER has achieved excellent performance, other frameworks are also being tried. Li et al.\cite{li-etal-2020-unified} utilized BERT to perform NER tasks in the MRC framework on general-domain datasets (e.g., the news dataset). Their BERT-MRC model obtains SOTA performance on both ``flat" and ``nested" datasets. In this work, we reproduced BERT-MRC (denoted by BERT-MRC (L)) on the six BioNER datasets. Specifically, we utilize the same weights and queries as BioBERT-MRC to perform BERT-MRC (L) for a fair comparison. Similar to BERT-MRC, we also formulate the BioNER task as an MRC problem, and employ BioBERT to implement BioNER tasks in the MRC framework. It is encouraging to see that BioBERT-MRC obtains SOTA performance on all the six datasets. Compared with feature engineering-based and LSTM-CRF-based methods, our method usually obtains both the highest precision and recall performance. This shows that our method is far superior to those methods. Compared with BioBERT-Softmax, BioBERT-MRC can greatly improve the precision, thereby increasing the F1-score. This experimental result indicates that our method is also superior to BioBERT-Softmax. Furthermore, the performance of BioBERT-MRC is also better than that of BERT-MRC (L). The main difference between BioBERT-MRC and BERT-MRC is the loss function. The loss function of BERT-MRC contains three parts, namely $\alpha Loss_{start}$ + $\beta Loss_{end}$ + $\gamma Loss_{span}$ ($\alpha, \beta, \gamma \in [0,1]$ are hyper-parameters), but the function of BioBERT-MRC has only two parts, i.e., ($Loss_{start}$ + $Loss_{end}$)$/2$. BERT-MRC is designed for all types of NER entities (including ``flat" and ``nested" entities), but all the six BioNER datasets BERT-MRC (L) run only contain the ``flat" entities. This indicates that $Loss_{span}$ designed for ``nested" entities is meaningless, and it may be the main reason why the performance of BERT-MRC is weaker than BioBERT-MRC. In conclusion, these experimental results demonstrate that BioBERT-MRC is superior in performance as compared with other existing methods.

\subsection{Case Study}
To further explore the difference between BioBERT-MRC and BioBERT-Softmax, we conducted the case study. The results of case study are shown in Table VIII. For the first example, this is a case where BioBERT-Softmax recognized false negatives (FNs) while BioBERT-MRC corrected them as true positives (TPs). This example shows that BioBERT-MRC has certain advantages over BioBERT-Softmax in terms of learning syntactic information (e.g., phrases and segments). The second example is a consistency problem. It can be seen that BioBERT-Softmax only recognized one ``cocaine" in the whole sequence, while BioBERT-MRC corrected the error of BioBERT-Softmax. This example shows that BioBERT-MRC can alleviate the problem of label inconsistency by learning the semantic information of the entire sequence. The third and last examples are a case where BioBERT-Softmax recognized false positives (FPs) while BioBERT-MRC corrected them as true negatives (TNs). The fourth and fifth are segmentation problem examples. Compared with BioBERT-Softmax, BioBERT-MRC can better distinguish the boundary information of entities. These four examples all demonstrate the effectiveness of BioBERT-MRC in the syntactic learning. Through the case study, we can infer that compared with BioBERT-Softmax, BioBERT-MRC has better performance in syntactic and semantic learning. Specifically, BioBERT-MRC can correct some FNs and FPs, accurately recognize entity boundaries, and alleviate the label inconsistency problem.

\section{Conclusion}
In this work, we use BERT to perform BioNER in the MRC framework. Compared with using BERT in the sequence labeling framework, performing BERT (i.e., BioBERT) in the MRC framework can enhance the model’s ability to recognize target entities. Moreover, the MRC framework essentially has the advantage of introducing prior knowledge, and the performance of BioBERT-MRC can be effectively improved on most datasets by designing queries. The proposed approach achieves SOTA performance on six BioNER datasets, which demonstrates its effectiveness. In future work, we would like to explore the effectiveness of transfer learning in the MRC framework.

\section*{Conflict of interest}
The authors declare that they have no known competing financial interests or personal relationships that could have appeared to influence the work reported in this paper.

\section*{Acknowledgment}
This work was supported by the National Key Research and Development Program of China under Grant 2016YFC0901902.

\bibliographystyle{IEEEtran}
\end{document}